\documentclass[conference]{IEEEtran}
\IEEEoverridecommandlockouts

\usepackage{cite}
\usepackage{amsmath,amssymb,amsfonts}
\usepackage{algorithmic}
\usepackage{graphicx}
\usepackage{textcomp}
\usepackage{xcolor}
\usepackage{multirow}
\usepackage{jabbrv}
\usepackage{tablefootnote}
\usepackage{threeparttable}

\begin{document}


\title{Towards Scalable Handwriting Communication via EEG Decoding and Latent Embedding Integration\\ 
\thanks{This work was partly supported by Institute of Information \& Communications Technology Planning \& Evaluation (IITP) grant funded by the Korea government (MSIT) (No. RS--2021--II--212068, Artificial Intelligence Innovation Hub, No. RS--2024--00336673, AI Technology for Interactive Communication of Language Impaired Individuals, and No. RS--2019--II190079, Artificial Intelligence Graduate School Program (Korea University)).}
}

\author{\IEEEauthorblockN{Jun-Young Kim}
\IEEEauthorblockA{\textit{Dept. of Artificial Intelligence} \\
\textit{Korea University}\\
Seoul, Republic of Korea \\
j\_y\_kim@korea.ac.kr}
\and
\IEEEauthorblockN{Deok-Seon Kim}
\IEEEauthorblockA{\textit{Dept. of Artificial Intelligence} \\
\textit{Korea University}\\
Seoul, Republic of Korea \\
deokseon\_kim@korea.ac.kr}
\and
\IEEEauthorblockN{Seo-Hyun Lee}
\IEEEauthorblockA{\textit{Dept. of Brain and Cognitive Engineering} \\
\textit{Korea University}\\
Seoul, Republic of Korea \\
seohyunlee@korea.ac.kr}
}

\maketitle

\begin{abstract}
In recent years, brain--computer interfaces have made advances in decoding various motor--related tasks, including gesture recognition and movement classification, utilizing electroencephalogram (EEG) data. These developments are fundamental in exploring how neural signals can be interpreted to recognize specific physical actions. This study centers on a written alphabet classification task, where we aim to decode EEG signals associated with handwriting. To achieve this, we incorporate hand kinematics to guide the extraction of the consistent embeddings from high--dimensional neural recordings using auxiliary variables (CEBRA). These CEBRA embeddings, along with the EEG, are processed by a parallel convolutional neural network model that extracts features from both data sources simultaneously. The model classifies nine different handwritten characters, including symbols such as exclamation marks and commas, within the alphabet. We evaluate the model using a quantitative five--fold cross--validation approach and explore the structure of the embedding space through visualizations. Our approach achieves a classification accuracy of 91~\% for the nine--class task, demonstrating the feasibility of fine--grained handwriting decoding from EEG.

\end{abstract}


\section{INTRODUCTION}

Over the past few decades, brain--computer interfaces (BCIs) have been developed across various fields to enhance human convenience\cite{kim2015abstract}. Particularly, the electroencephalogram (EEG) has enabled non--invasive data collection, offering a balance between signal--to--noise ratio (SNR) and ease of acquisition, which has driven extensive research. For instance, studies have decoded sleep stages to monitor rest quality\cite{diykh2016eeg}, mental states to prevent cognitive fatigue--related accidents\cite{han2020classification, jeong2019classification, guan2022eeg, ahn2022multiscale, lee2020continuous}, motor imagery for controlling external devices\cite{kilteni2018motor , mane2020multi}, imagined speech for communication efficiency\cite{willett2023high}, as well as classified schizophrenia from the EEG to aid in diagnosis and treatment\cite{prabhakar2020framework}. These works focus on enhancing seamless interaction and effective communication through BCI systems.

Research on decoding motor intentions has progressed in aiding motor-impaired individuals by developing EEG-based control systems. The EEG-based robotic arm control for tetraplegics achieved 70.5\% accuracy in feedback training\cite{onose2012feasibility}. A convolutional neural network (CNN) followed by a bidirectional long short--term memory network for 3--D arm movement decoding showed approximately 60\% success in real-time robotic control\cite{jeong2020brain}. However, these studies are often limited to broader motor decoding, which can pose challenges in achieving finer motor detail decoding. Fine motor decoding in particular requires highly detailed neural information, yet EEG signals are often limited by their high dimensionality and noise, complicating efforts to achieve precise classification. Such constraints highlight the need for methods capable of capturing meaningful, low-dimensional embeddings from complex neural data.


In speech decoding, low and high--frequency neural activity has been shown to play a significant role in imagined speech decoding, particularly in phonetic spaces \cite{proix2022imagined}. The high--frequency EEG features have demonstrated superior classification accuracy in imagined speech and visual imagery \cite{lee2020neural}. Recently, the Diff--E model, using denoising diffusion probabilistic models (DDPM), significantly improved imagined speech decoding accuracy \cite{kim2023diff}. Despite these advances, EEG-based imagined speech decoding is still challenged by issues related to the signal-to-noise ratio (SNR).

\begin{figure*}[htb!]
  \centering
  \includegraphics[width=0.88\linewidth,height=\textheight,keepaspectratio]{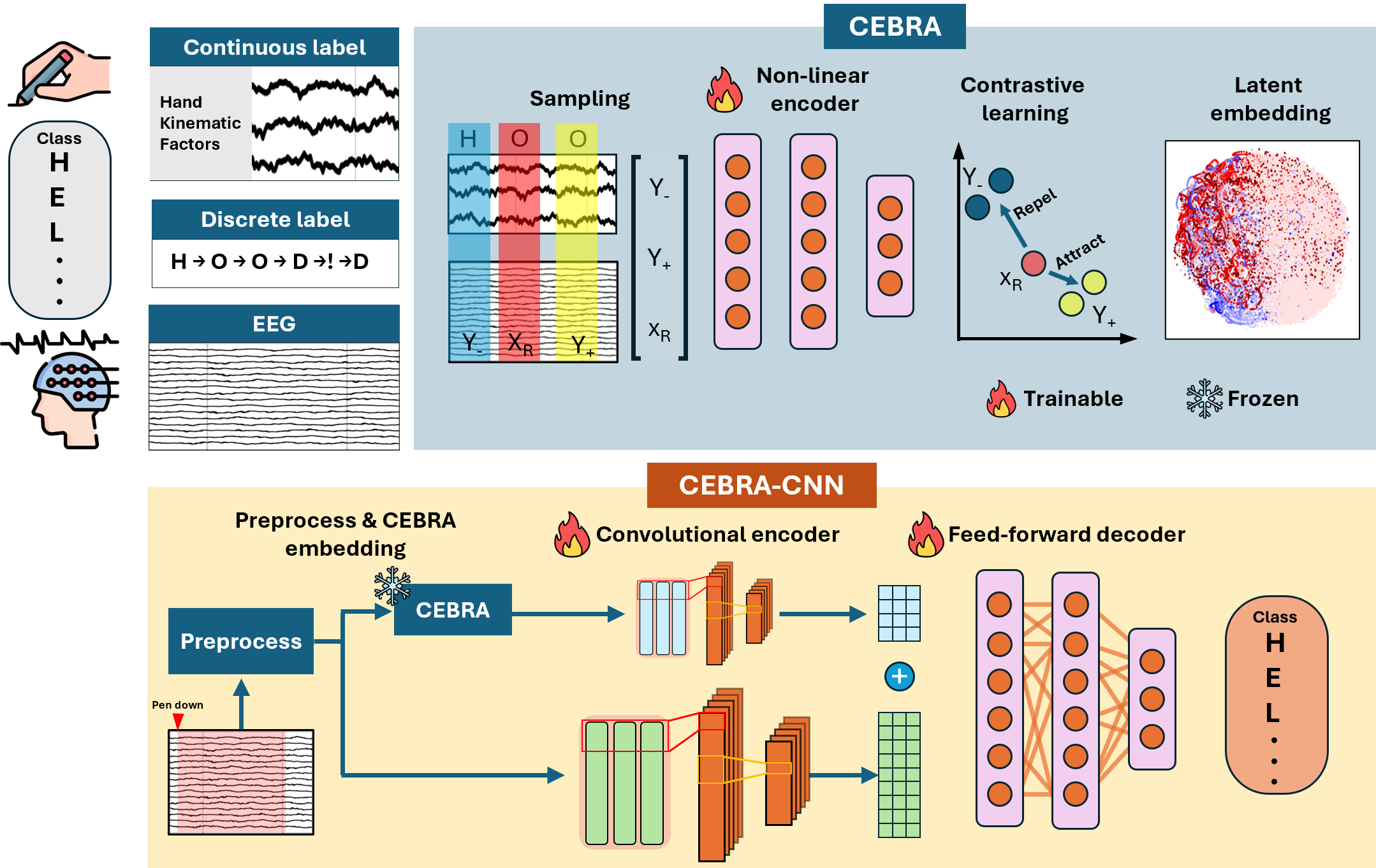}
  \caption{Framework for hand kinematics--aware consistent embedding generation and feature fusion using CEBRA--CNN.}
  \label{fig:method}
\end{figure*}


To address the limitations of traditional EEG decoding in capturing fine motor details and handling low signal--to--noise ratios in imagined speech tasks, we employ a recently proposed encoding method, consistent embeddings from high--dimensional neural recordings using auxiliary variables (CEBRA) \cite{schneider2023learnable}, which enables robust and consistent latent representations through contrastive learning. CEBRA is designed to jointly utilize behavioral and neural data, generating interpretable embeddings that maintain consistency across sessions, tasks, and subjects. By leveraging auxiliary variables such as time or behavioral context, CEBRA efficiently maps neural dynamics, creating reliable low--dimensional embeddings even for complex, high--dimensional data. CEBRA has demonstrated high performance in generating behaviorally relevant embeddings from invasive animal recordings, such as Neuropixels and two--photon calcium imaging. However, CEBRA has not yet been tested on non--invasive human EEG data. In this study, we apply CEBRA to EEG signals collected during a written alphabet classification task, exploring its potential to generate consistent embeddings in this domain. A CNN--based model jointly processes the EEG data and the CEBRA embeddings, achieving a written alphabet recognition accuracy of 92.1\%, which outperforms traditional classification models such as CNN, EEGNet\cite{lawhern2018eegnet}, and DeepConvNet\cite{schirrmeister2017deep}.


\section{MATERIALS AND METHODS}
    \subsection{Preparation of EEG and Kinematic Data}
    We used the handwriting kinematics--EEG dataset \cite{pei2021online}. The experimental paradigm involved participants wearing the EEG electrodes using 32--channel amplifier (BrainAmp, Brain Products GmbH, Germany) while repetitively performing a handwriting task on a fixed tablet PC (HUAWEI MatePad Pro) display. The participants required the to repeatedly write the phrase “HELLO, WORLD!” including special characters, for a total of 300 trials. During this process, both hand kinematics (i.e., $x$, $y$, $pressure$, $velocity$) and the EEG were simultaneously recorded. Pen--down and pen--up triggers were collected and these triggers were used to align the EEG data with the handwriting events. The EEG was recorded at a sampling rate of 250 Hz, while the tablet captured handwriting trajectories in real--time.
    
    For preprocessing, the EEG data was first re--referenced using the average reference. A bandpass filter between 1 to 45 Hz was applied to remove unwanted frequency components, and independent component analysis was performed to remove eye movement artifacts. Subsequently, a second low--pass filter from 0.5 to 8 Hz was applied, based on the findings in the existing literature \cite{pei2021online}, which indicate that handwriting--related EEG features are most prominent in the lower frequency bands. Epoching was then conducted based on the pen--down event, extracting 0 to 1 sec. time windows of the EEG data relative to the trigger. 
    
    And then we extracted the onset trajectory for each character and interpolated it 0 to 1 sec. time windows to align it with the EEG data. Both the EEG and trajectory data were normalized before being stored in a 5--fold cross--validation scheme. EEG data was z--normalized per channel, while the handwriting trajectories were min--max normalized to reduce distortions caused by scaling differences. Additionally, the starting point of each trial was shifted to the origin to maintain consistency across trials.
    
    \subsection{Joint Learning: CEBRA Embeddings and EEG Data}

    Hand kinematics are integrated into a two--stage learning process. In the first stage, the CEBRA embeddings are learned from EEG to capture the underlying structure of the neural dynamics related to fine motor behavior. In the second stage, the model fuses features from both the preprocessed EEG and the learned CEBRA embeddings through a parallel CNN architecture, enhancing classification performance.

        \subsubsection{Learning CEBRA embeddings}
        As shown in Fig.~\ref{fig:method}, the first stage of our method involves training the CEBRA model to extract meaningful low--dimensional latent representations from the high--dimensional EEG.

        For each fold of the dataset, we load the preprocessed EEG and corresponding continuous and discrete labels. The preprocessed EEG undergoes transformation through the CEBRA, parameterized by the model’s hyperparameters, such as a batch size of 1,024, learning rate of 0.0002, and a temporal offset window of 10 frames. The model is trained to map the neural signals into a lower--dimensional embedding space of size $D_{embed}$, which varies across different experimental configurations (i.e., $D_{embed} = \text{2, 4, 8, 12, and 16}$). The trained model produces embeddings $\mathcal{E} \in \mathbb{R}^{N \times D_{embed}}$ for each dataset, where $N$ is the total number of time points. These embeddings are then saved for use in the classification task.
        
        \subsubsection{CNN--based feature fusion}
        After obtaining the CEBRA embeddings, we design a classification model that jointly leverages both the preprocessed EEG and the learned embeddings. The $\mathcal{D}_{EEG}$ is processed through a CNN designed to capture spatial and temporal dependencies within the signal. Simultaneously, the CEBRA embeddings $\mathcal{E}$ are passed through a parallel CNN designed to process the low--dimensional data.
        
    \begin{table}[t!]
    \renewcommand{\arraystretch}{1.25}
    \centering
    \begin{threeparttable}
    \caption{Comparison of Classification Performance (\%) Across Different Models and the CEBRA Embedding Dimensions.}
    \begin{tabular}{lccc}
    \hline
    \textbf{Model} & \textbf{Acc. ± Std.} & \textbf{F1--score ± Std.} \\ \hline
    \textbf{DeepConvNet \cite{schirrmeister2017deep}}  & 73.0 ± 1.2  & 68.9 ± 3.6 \\
    \textbf{EEGNet \cite{lawhern2018eegnet}}  & 84.8 ± 1.1  & 84.2 ± 0.5 \\ 
    \textbf{CNN}  & 91.1 ± 1.6  & 90.8 ± 2.6 \\ 
    \textbf{CNN + 2--D CEBRA Embeddings}  & 89.6 ± 1.3  & 90.7 ± 2.0 \\ 
    \textbf{CNN + 4--D CEBRA Embeddings}  & 91.1 ± 1.5  & 90.2 ± 1.6 \\ 
    \textbf{CNN + 8--D CEBRA Embeddings}  & 91.3 ± 1.1  & 90.8 ± 1.9 \\ 
    \textbf{CNN + 12--D CEBRA Embeddings} & 91.4 ± 1.3  & 90.7 ± 1.4 \\ 
    \textbf{CNN + 16--D CEBRA Embeddings} & \textbf{92.1 ± 1.3}  & \textbf{91.5 ± 1.8} \\ \hline
    \end{tabular}
    \begin{tablenotes}
        \footnotesize
        \item \textbf{Acc.}: accuracy; \textbf{Std.}: standard deviation.
    \end{tablenotes}
    \label{table:performance_metrics}
    \end{threeparttable}
    \end{table}
        The architecture of the joint model consists of two separate CNN branches: one for processing the raw neural signals and the other for processing the embeddings. The outputs of both branches are concatenated, and the combined representation is fed into a series of fully connected layers to produce the final classification output. The classifier is trained using a cross--entropy loss function, and training is performed using an Adam optimizer with a learning rate of 0.001. This joint model allows us to combine the benefits of the high--dimensional raw data and the structured, low--dimensional CEBRA embeddings, which leads to improved robust classification performance.
    
        \subsubsection{Evaluation and visualization}
        The model is trained and evaluated using a 5--fold cross--validation scheme. For each fold, we report the classification accuracy on a held--out validation set. To demonstrate the effectiveness of the CEBRA embeddings, we apply dimensionality reduction techniques, including \textit{t}--distributed stochastic neighbor embedding (\textit{t}--SNE) and principal component analysis (PCA), on the CEBRA embeddings and EEG data, enabling a qualitative assessment of class separability within the embedding~space.\\

\section{RESULTS AND DISCUSSION}
    \subsection{EEG Decoding Performance}

    \begin{figure}[t]
      \centering
      \includegraphics[width=0.92\linewidth,height=\textheight,keepaspectratio]{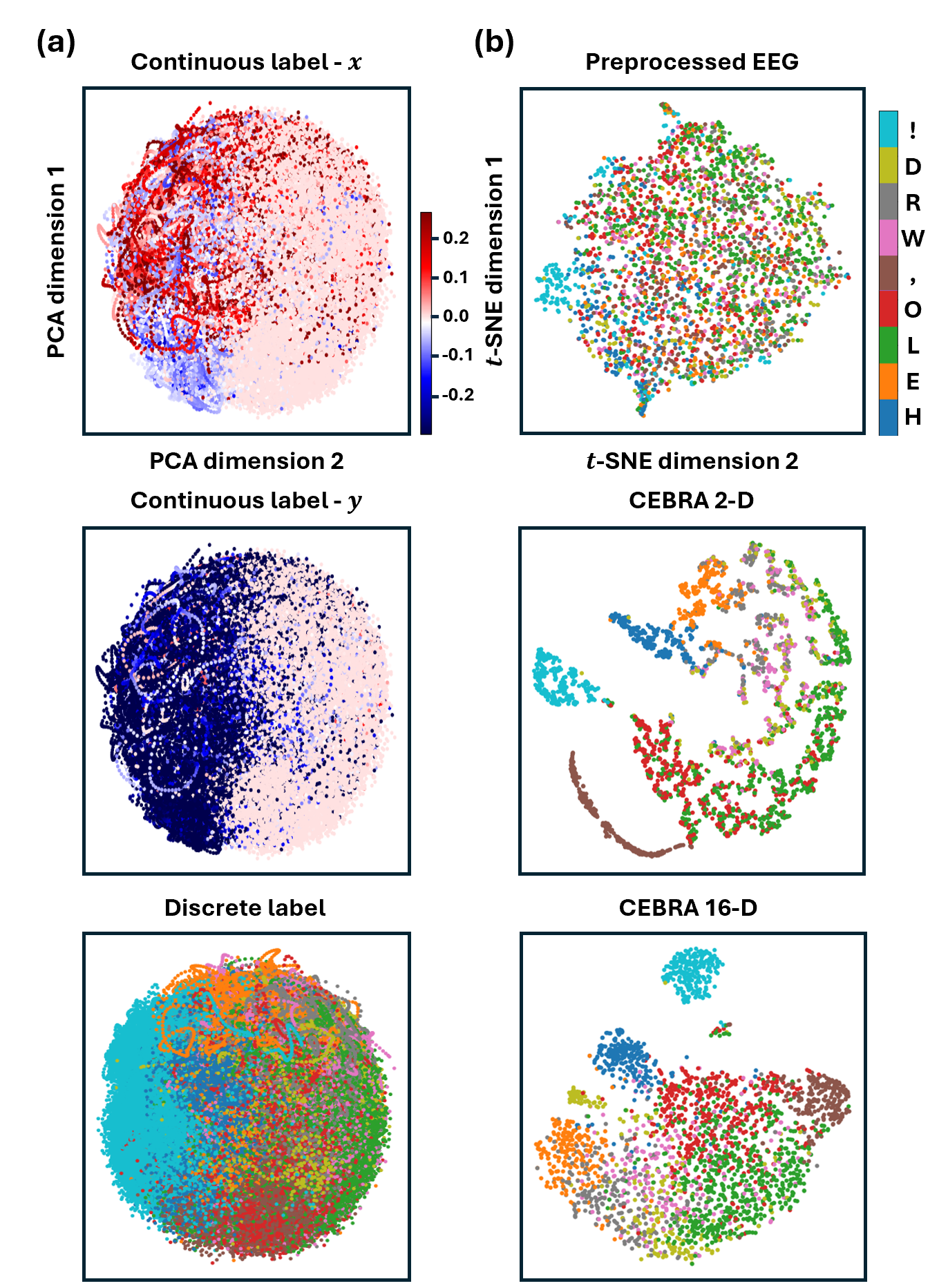}
      \caption{(a) PCA projection of the CEBRA embeddings, showing the structure of both continuous and discrete labels. (b) \textit{t}--SNE visualization comparing the clustering of the preprocessed EEG against the 2--D and 16--D CEBRA embeddings.}
      \label{fig:visualization}
    \end{figure}
    
    Table \ref{table:performance_metrics} summarizes the classification performance across different models and embedding dimensions. Among all models, the CNN with CEBRA 16--D embeddings achieved the best performance, with an accuracy of 92.1~\% and an F1--score of 91.5~\%. This highlights the advantage of using higher--dimensional embeddings, as they capture more detailed neural dynamics, leading to superior decoding accuracy.
    
    As the embedding dimensions decreased, performance gradually declined, with the CEBRA 2--D embeddings still maintaining a competitive accuracy of 89.6~\%. On the other hand, conventional models without any embedding integration, such as DeepConvNet\cite{schirrmeister2017deep} and EEGNet\cite{lawhern2018eegnet}, exhibited lower performance, with accuracies of 84.8~\% and 73.0~\%, respectively. Interestingly, the baseline CNN model, which uses only preprocessed EEG data without any CEBRA embeddings, achieved a competitive accuracy of 91.1~\%, similar to several CEBRA--CNN models. However, the CEBRA--CNN model with 16--D embeddings still outperformed the baseline CNN, demonstrating the advantages of integrating high--dimensional latent embeddings for enhanced classification accuracy.

    \subsection{Data Embeddings}
    As shown in Fig.~\ref{fig:visualization}(a), the embeddings capture hand kinematics with clear separation along $x$ and $y$ coordinates and well--formed clusters under discrete class labels, suggesting that task features are generally well--preserved. However, for continuous labels, some embeddings were incorrectly mapped near the origin, indicating potential for improvement in handling continuous features. As a potential improvement, incorporating diffusion models could iteratively denoise and refine these embeddings, enhancing SNR and enabling more precise, robust embeddings that support a wider range of EEG classification tasks. The fundamental philosophy of DDPM lies in a divide--and--conquer approach, allowing high--quality data synthesis even from pure noise without discernible distribution patterns. This capability could be especially beneficial for low SNR EEG data by gradually restoring latent signal structures through denoising, thus improving the clarity and reliability of neural embeddings in complex decoding tasks\cite{kim2023diff}.

    
    As shown in Fig.~\ref{fig:visualization}(b), \textit{t}--SNE was applied to both the raw EEG data~and~the~2--D and 16--D CEBRA embeddings. Although both embeddings demonstrated distinct class clustering, the 16--D embeddings more effectively preserved the global structure, resulting in superior classification performance. This improvement in accuracy is likely attributed to the enhanced generalization achieved in the higher-dimensional space. Conversely, while the 2--D embeddings better captured the local structure, they lacked the refined separation provided by the higher--dimensional representation.
    
    In summary, the CEBRA--CNN models with higher--dimensional embeddings outperform conventional models like EEGNet and DeepConvNet, offering more accurate and robust decoding for EEG--based BCI systems.\\
    
\section{CONCLUSION}

In this study, we applied CEBRA to EEG data to generate consistent embeddings that improve fine motor task classification, specifically for written alphabet recognition, by integrating CEBRA embeddings with EEG data in a CEBRA-CNN architecture. Our approach achieved a classification accuracy of 92.1\% on a 9--class dataset, demonstrating the effectiveness of combining high-dimensional CEBRA embeddings with traditional EEG processing for improved class separability and neural dynamics representation. The 16-D embeddings, in particular, retained essential motor features, as visualized through \textit{t}--SNE and PCA, highlighting clearer neural structure and enhanced class separability. Compared to baseline models, the addition of CEBRA embeddings led to improved robustness in classification, particularly for EEG--based fine motor tasks. This study underscores the potential of CEBRA--augmented embedding spaces for advancing EEG-based BCI applications, paving the way for more precise and adaptable communication systems. Future work could explore refining these embeddings to further address signal-to-noise limitations and enhance classification precision.


\bibliographystyle{jabbrv_IEEEtran}
\bibliography{REFERENCE}


\end{document}